\documentclass[cameraready]{Interspeech}
\usepackage{amssymb}

\title{S-DiverSe: Spanish Diverse Speech}

\author[affiliation={1, 2}, orcid=0000-0002-7705-2250, equalcontribution, correspondingauthor]{Fernando}{López}
\author[affiliation={2}, equalcontribution]{Fernando}{Ibañez}
\author[affiliation={2}]{Ana}{Martínez}
\author[affiliation={2}]{Iván}{Alonso}
\author[affiliation={1}]{\\Pablo}{Gómez}
\author[affiliation={3}, orcid=0000-0002-3725-742X,]{Santosh}{Kesiraju}
\author[affiliation={1}, orcid=0000-0002-4507-4930,]{Jordi}{Luque}

\address{
    $^1$ Scientific Research, Telefónica Innovación Digital, Spain \\
    $^2$ Universidad Autónoma de Madrid, Spain \\
    $^3$ Brno University of Technology, Czech Republic
}

\email{fernando.lopez@telefonica.com}


\usepackage{comment}

\begin{document}

\maketitle

\begin{abstract}
    Automatic speech recognition (ASR) has advanced remarkably for standard speech, yet speech affected by neurological conditions remains a challenge. We present S-DiverSe (Spanish Diverse Speech), a corpus of 3.2 hours of in-the-wild Spanish speech from 22 speakers with amyotrophic lateral sclerosis, Parkinson's disease, and stroke. The dataset contains 444 manually transcribed audio segments with metadata on speaker sex, disease type, and intelligibility. S-DiverSe is designed to support ASR evaluation and development for neurologically affected Spanish speech. We describe the dataset, analyze its composition, and report baseline ASR results alongside initial adaptation experiments. Our findings reveal that heuristic text post-processing is more robust than fine-tuning for out-of-domain neurological Spanish speech. This underscores the need for dedicated in-the-wild Spanish benchmarks.
\end{abstract}

\section{Introduction}
\label{sec:intro}
Automatic speech recognition (ASR) has improved markedly in recent years, with state-of-the-art models achieving low word error rates on standard benchmarks \cite{ahlawat2025automatic}. However, they continue to face substantial challenges when applied to real-world speech scenarios \cite{maheshwari2025asr}. One of these challenges is the recognition of speech produced by individuals with neuromotor disorders, such as amyotrophic lateral sclerosis (ALS), Parkinson's disease (PD), and post-stroke conditions. These conditions frequently cause dysarthria, which impairs neuromuscular control of speech and leads to reduced articulation clarity, altered prosody, and variable intelligibility, all of which challenge ASR systems \cite{rudzicz2012torgo, kim2008dysarthric}.

A central obstacle in pathological speech recognition is the lack of large, high-quality datasets \cite{hasegawa2024community}. This remains true even for English, despite resources such as UA-Speech \cite{kim2008dysarthric}, TORGO \cite{rudzicz2012torgo}, and the recent Interspeech Speech Accessibility Project (SAP) challenge \cite{hasegawainterspeech}. For Spanish, the gap is wider: publicly available corpora for neurological conditions are scarce and typically limited in either clinical coverage or domain diversity.

The Chilean Spanish dataset from the GITA laboratory \cite{orozco2014new} is a significant early contribution. It encompasses phonation tasks such as isolated vowels, vowel sequences, and changing tones, alongside diadochokinetic evaluations, word and sentence repetitions, and spontaneous speech. However, it lacks diversity in recording conditions and is predominantly biased towards an elderly demographic. More recently, NeuroVoz \cite{mendes2024neurovoz, mendes_laureano_2024_10777657} has provided Castilian Spanish data for PD collected under controlled hospital conditions, including sustained vowels, sentence repetitions, diadochokinetic evaluation, and short monologues. While valuable, it primarily contains brief, elicited utterances from elderly participants, limiting its ability to capture the broader variability of neurologically affected speech.

Consequently, rigorous evaluation of Spanish ASR for speech affected by different neurological conditions remains difficult in the absence of benchmarks. This limitation hinders both the development of robust ASR for pathological speech and progress on assistive communication and clinical assessment tools. To bridge this gap, we make three contributions:
(i) We introduce S-DiverSe (\textbf{S}panish \textbf{Diver}se  \textbf{S}p\textbf{e}ech), the first in-the-wild Spanish dataset covering multiple neurological conditions. It includes human transcripts and metadata for sex, condition, and intelligibility, and we release annotations to foster open research\footnote{We distribute exclusively annotations and video links: \url{https://github.com/ferugit/s-diverse}}. (ii) We evaluate four state-of-the-art ASR systems on S-DiverSe and compare performance against other neuro-affected datasets. (iii) We use S-DiverSe as a neuro-affected speech benchmark, and evaluate post-processing and fine-tuning techniques for two ASR models to quantify the benefits and limits of adaptation for Spanish neurological speech.

\section{S-DiverSe: Spanish Diverse Speech}
S-DiverSe is designated for speech recognition tasks. It comprises 3.2 hours of human-transcribed speech, extracted from in-the-wild recordings. It features 22 unique speakers with ALS, PD, and post-stroke aftereffects. The corpus contains 444 audio segments of variable duration, and provides metadata for speaker sex, condition, and intelligibility.

\begin{table}[h!]
\centering
\caption{Sex and pathological condition duration on S-DiverSe.}
\label{tab:dataset_summary}
\resizebox{\columnwidth}{!}{
\begin{tabular}{l cc ccc}
\toprule
& \multicolumn{2}{c}{\textbf{Sex}} & \multicolumn{3}{c}{\textbf{Condition}} \\
\cmidrule(lr){2-3} \cmidrule(lr){4-6}
\textbf{Metric} & \textbf{Male} & \textbf{Female} & \textbf{ALS} & \textbf{PD} & \textbf{STROKE} \\
\midrule
Duration (h) & 2.80 & 0.41 & 2.51 & 0.52 & 0.18 \\
Percentage & 87.4\% & 12.6\% & 78.1\% & 16.1\% & 5.8\% \\
\bottomrule
\end{tabular}
}
\end{table}

\subsection{Statistics}
\label{sec:stats}
\begin{figure}[h!]
    \centering
    \includegraphics[width=1.0\linewidth]{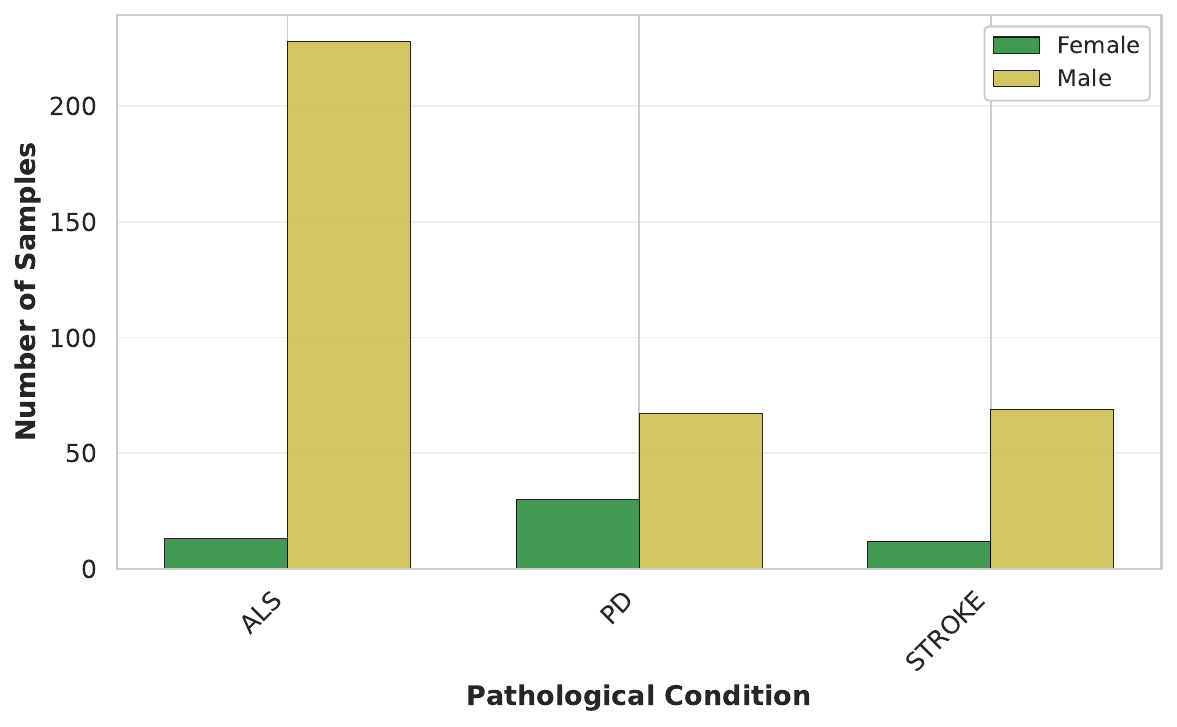}
    \vspace{-0.7cm}
    \caption{Sex distribution by pathological condition.}
    \label{fig:sex_and_speech_condition}
\end{figure}

Table~\ref{tab:dataset_summary} reports total duration in hours per sex and pathological condition. The corpus is male-dominant and ALS-dominant. Sex and pathology imbalance mirrors the availability of what is findable in-the-wild rather than a collection choice. Figure~\ref{fig:sex_and_speech_condition} depicts the sample count by sex and per pathology.

\begin{figure}[h!]
    \centering
    \includegraphics[width=1.0\linewidth]{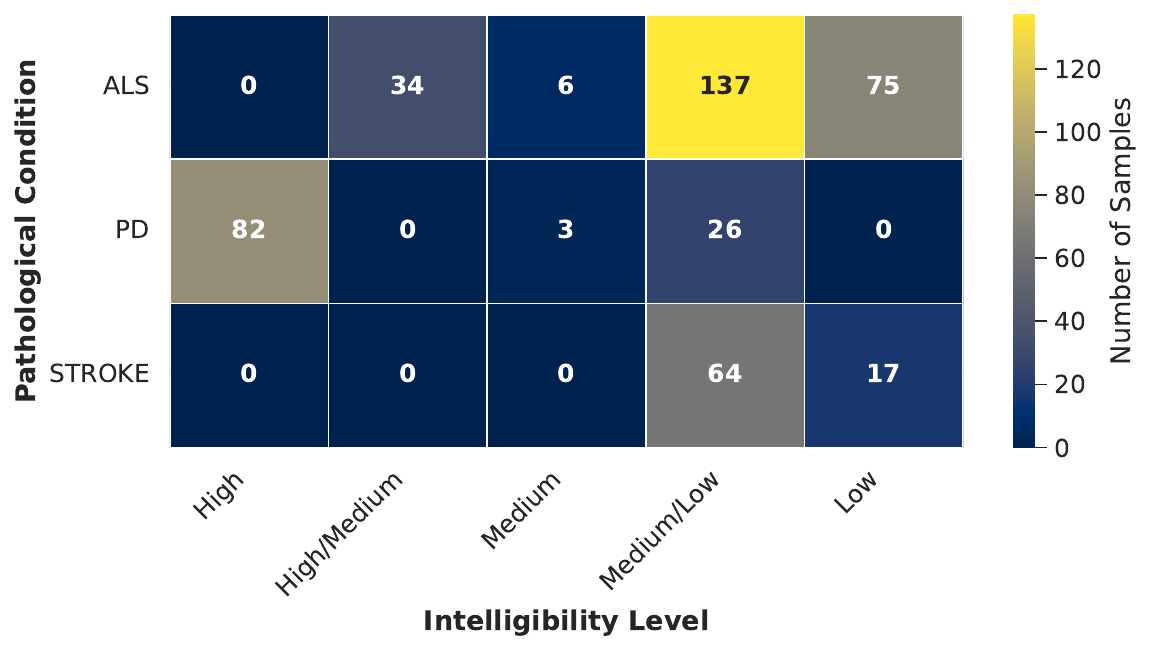}
    \vspace{-0.7cm}
    \caption{Sample distribution of condition vs intelligibility.}
    \label{fig:condition_vs_inteligibility}
\end{figure}

Figure~\ref{fig:condition_vs_inteligibility} shows the distribution of samples by condition and intelligibility. ALS and stroke exhibit unimodal distributions skewed toward medium-to-low intelligibility, while PD displays a bimodal pattern.

\begin{figure}[h!]
    \centering
    \includegraphics[width=1.0\linewidth]{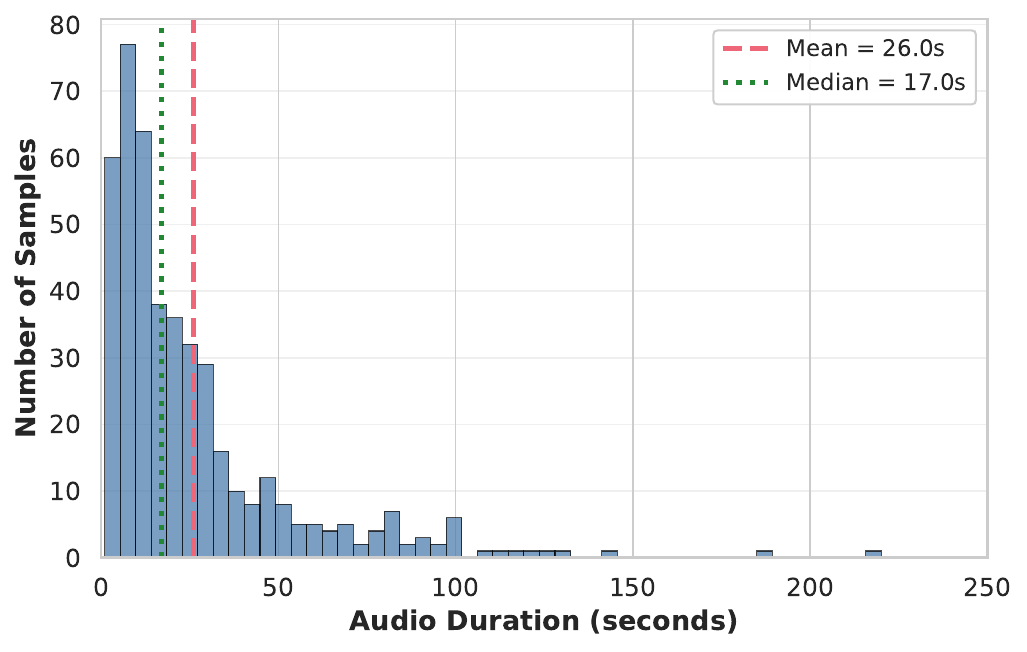}
    \vspace{-0.7cm}
    \caption{Audio Duration Distribution.}
    \label{fig:length_distribution}
\end{figure}

Segment durations are predominantly under 30 seconds, with sparse coverage beyond 50 seconds and few recordings reaching approximately 250 seconds (Figure~\ref{fig:length_distribution}).

Transcript linguistic complexity was estimated using \texttt{salamandra-2b} \cite{gonzalez2025salamandra}, a Castilian Spanish open-weight LLM, yielding a median perplexity of 33, indicating moderate linguistic simplicity.  As a reference point, we computed the median perplexity of the Spanish Common Voice v24.0 \cite{ardila2020common} training set, which yielded 49. 

\subsection{Source and annotation}
Audio was sourced from in-the-wild YouTube recordings. Three native Spanish-speaking Linguistics graduates identified candidate videos, filtered content, segmented recordings, and produced transcriptions.

Videos were retrieved using Spanish queries targeting dysarthric speech (e.g., condition-specific keywords, interviews, and documentaries), with a preference for spontaneous speech contexts such as interviews. This yields acoustically diverse conditions, including background noise, music, and heterogeneous recording setups. Candidates were filtered based on self-reported diagnoses, video metadata, and perceptual evidence of non-normative speech.

Segments were defined by speaker turns and discourse coherence, with variable durations to preserve semantic unity. Each segment was assigned to a single annotator, who produced orthographic transcriptions, labeled speaker sex and condition, and assigned an intelligibility score. Unintelligible segments were transcribed as \texttt{<unk>}, and common filled pauses (e.g., “mmm”, “eh”, “em”) were retained. Transcriptions were then cross-reviewed across annotators.

Intelligibility was rated on a 1--5 ordinal scale (1: low intelligibility; 5: high intelligibility), based on perceived transcription effort, including the presence of unintelligible spans within a turn. Inter-annotation agreement (IAA) was assessed on a stratified subset of 50 samples covering all pathologies and intelligibility levels. Weighted Cohen's $\kappa$ (linear) was 0.38 (fair), with 74\% adjacent agreement ($\pm$1 level): most disagreements fell between neighboring categories, particularly at the ``Medium/Low"--``Medium" boundary. We therefore keep the original intelligibility annotation.

\section{Experimental setup}
S-DiverSe is used exclusively for ASR evaluation. The factor preventing further partitioning is not its modest size but its variability: it spans three pathologies, multiple intelligibility levels, and heterogeneous acoustic conditions, so any training split would be too skewed per stratum for reliable adaptation. Following the adaptation strategies explored in the SAP challenge \cite{hasegawainterspeech}, we investigate text post-processing and fine-tuning. Fine-tuning is done with data from different languages or the same language but a constrained domain, assessing cross-lingual transfer and domain mismatch effects.

\subsection{Models}
We evaluate three open-weight and one commercial ASR system, and additionally benchmark the open-weight models on other neuro-affected corpora.

\textbf{Whisper-large-v3} \cite{radford2023robust} is a 1.6B-parameter encoder-decoder model pre-trained on 1M hours of weakly labeled and 4M hours of pseudo-labeled audio.

\textbf{Voxtral-Mini} \cite{liu2025voxtral} is a 4.7B-parameter model integrating a fine-tuned Whisper-large-v3 encoder connected to a fine-tuned Ministral-3B LLM \cite{liu2026ministral3} via a connector. We used the \texttt{Voxtral-Mini-3B-2507} checkpoint.

\textbf{omniASR\_CTC\_1B\_v2} \cite{omnilingual2025omnilingual} features a wav2vec2.0-based transformer encoder \cite{baevski2020wav2vec} with nearly 1B parameters. It was first pre-trained via self-supervised learning on 4.3M hours of unlabelled data spanning over 1,600 languages, then fine-tuned for ASR using a CTC head.

\textbf{ElevenLabs Scribe v2} \cite{elevenlabs_scribe_v2} is a commercial black-box baseline, accessed via cloud API with the \texttt{scribe-v2} model under identical settings across all test conditions (January 2026).

\subsection{Databases}
\label{sec:databases}
To broaden our analysis, TORGO and NeuroVoz are used both for cross-dataset evaluation and as adaptation training data.

\textbf{TORGO} is an English dysarthric speech corpus including speakers with cerebral palsy, covering diadochokinetic tasks, sustained phonation, isolated words, and constrained and unconstrained sentences. We retain the isolated-word and constrained-sentence subsets, yielding 13.68 hours ($\sim$16.6k utterances) from 15 speakers: 3 female dysarthric, 3 female HC, 5 male dysarthric, and 4 male HC. The subset comprises 8.66 hours of male and 5.02 hours of female speech, with 7.85 hours of single-word and 5.83 hours of multi-word utterances.

From \textbf{NeuroVoz} \cite{mendes2024neurovoz}, introduced in Section~\ref{sec:intro}, we retained sentence repetitions and short monologues, yielding 2.31 hours (1.8k samples) from 111 speakers: 26 female and 28 male HCs, and 20 female and 33 male PD speakers, with 4 of unknown sex. Duration totals 1.28 hours (male), 0.95 hours (female), and the remainder from unknown-sex speakers.

Both datasets are split into train/validation/test partitions at a 70/10/20 ratio, stratified by speaker identity to prevent leakage, with both sexes and pathological and healthy speech represented in every split.

To supplement target-language data, we use Spanish Common Voice v24.0 \cite{ardila2020common}, filtering 7.3 hours ($\sim$4.9k samples) of read speech from the original training set, split into train/validation subsets stratified by speaker.

\subsection{Training}
We adapt Whisper-large-v3 and Voxtral-Mini with two strategies: Low-Rank Adaptation (LoRA) and full fine-tuning. We experiment with: (i) FFT: full fine-tuning; (ii) F-LoRA: LoRA applied to all attention and feed-forward layers; (iii) EFT: audio encoder fine-tuning, plus connector for Voxtral-Mini; (iv) E-LoRA: LoRA restricted to the encoder (plus connector for Voxtral-Mini).

Hyperparameters were tailored per strategy: FFT used conservative settings  (lr=1e-5, 3 epochs) to mitigate catastrophic forgetting; EFT used moderate settings (lr=2e-5, 5 epochs); LoRA-based methods used more aggressive settings (lr=3e-4, 10 epochs) with r=8, $\alpha$=16, and dropout=0.1. All strategies shared an effective batch size of 16 via gradient accumulation, cosine learning rate scheduling with 100 warmup steps, and early stopping (patience=5) based on validation loss.

We evaluate three training data configurations: (i) TORGO and NeuroVoz combined, with NeuroVoz oversampled by a factor of 3 to compensate for the language imbalance; (ii) NeuroVoz only; and (iii) TORGO, NeuroVoz, and Spanish Common Voice, the latter added to mitigate language mismatch.

\subsection{Post-processing}
We evaluate rule-based post-processing (PP) to mitigate hallucinations, following \cite{tan2025cba}. Three sequential steps are applied: (i) words exceeding 15 characters are analyzed for internal character-level repetition and reduced to their base unit; (ii) consecutive repeated words are collapsed (word-level deduplication); (iii) consecutive repeated phrases are reduced to a single occurrence (phrase-level deduplication)

\subsection{Evaluation}
We report Word Error Rate (WER) on the TORGO and NeuroVoz test splits and on the full S-DiverSe set.

Total WER is computed by aggregating errors across all utterances, reflecting the dataset's sample distribution rather than the mean of subgroup WERs. References and hypotheses are normalized by lowercasing, removing punctuation, stripping \texttt{<unk>} tokens, and converting standalone digits to word forms. Filled pauses are retained as regular tokens and thus contribute to WER when present in the reference.

All experiments were run on two A100-SXM4-40GB GPUs using greedy decoding. Whisper-large-v3 and omniASR\_CTC\_1B\_v2 employ sliding-window inference with 30s and 35s windows, respectively, and a 5s overlap for audio exceeding model input limits.

\section{Results and discussion}
Table~\ref{tab:combined_wer_full} summarizes WER on NeuroVoz, TORGO, and S-DiverSe for all systems and adaptation settings.

\begin{table*}[h!]
\centering
\caption{WER (\%) on in-domain (NeuroVoz, TORGO) and out-of-domain (S-DiverSe) corpora. TORGO reports single/multi-word; S-DiverSe reports per-pathology WER. \textbf{Bold} indicates lowest and \underline{underline} second-lowest WER per column. Scribe v2 runs on S-DiverSe only due to cost and NeuroVoz license. FT: fine-tuning; NV: NeuroVoz, TG: TORGO, CV: CommonVoice.}
\small
\resizebox{\textwidth}{!}{%
\begin{tabular}{l c c c c c c c c c c c}
\toprule
\textbf{Model} & \textbf{Method} & \textbf{Train data} & \textbf{FT} & \textbf{NeuroVoz} & \multicolumn{3}{c}{\textbf{TORGO}} & \multicolumn{4}{c}{\textbf{S-DiverSe}} \\
\cmidrule(lr){6-8} \cmidrule(lr){9-12}
 & & & & & \textbf{Single-word} & \textbf{Multi-word} & \textbf{Total} & \textbf{ALS} & \textbf{PD} & \textbf{STROKE} & \textbf{Total} \\
\midrule
Voxtral-Mini     & None & None & $\times$ & 6.75 & 46.83 & 16.13 & 25.15 & 46.32 & 24.26 & 39.73 & 40.43 \\
Whisper-large-v3 & None & None & $\times$ & 19.03 & 39.71 & 13.02 & 20.86 & 38.16 & 24.88 & 92.61 & 36.43 \\
omniASR\_CTC\_1B\_v2 & None & None & $\times$ & 16.96 & 79.48 & 20.66 & 37.95 & 35.22 & 27.38 & 46.96 & 33.56 \\
\midrule
Scribe\_v2 & None & None & $\times$ & - & - & - & - & \textbf{20.64} & \textbf{19.51} & \textbf{31.69} & \textbf{20.69} \\
\midrule
Voxtral-Mini        & PP & None     & $\times$ & 6.87 & 46.33 & 12.00 & 22.09 & 22.96 & 23.91 & 39.73 & 23.73 \\
Whisper-large-v3    & PP & None     & $\times$ & 19.15 & 39.61 & 11.83 & 20.00 & \underline{21.99} & \underline{20.34} & \underline{36.28} & \underline{22.01} \\
\midrule
Voxtral-Mini        & FFT + PP      & NV+TG & $\checkmark$ & \underline{4.01} & \underline{19.57} & \underline{8.62} & \underline{11.84} & 27.36 & 23.01 & 45.57 & 26.81 \\
Whisper-large-v3    & FFT + PP      & NV+TG & $\checkmark$ & 20.55 & \textbf{18.97} & \textbf{7.37} & \textbf{10.78} & 147.20 & 62.58 & 156.23 & 125.68 \\
\midrule
Voxtral-Mini        & F-LoRA + PP   & NV+TG & $\checkmark$ & \textbf{3.53} & 25.04 & 12.17 & 15.96 & 43.10 & 36.31 & 48.03 & 41.51 \\
Whisper-large-v3    & F-LoRA + PP   & NV+TG & $\checkmark$ & 16.17 & 20.76 & 8.64 & 12.20 & 34.78 & 22.95 & 52.95 & 32.30 \\
\midrule
Voxtral-Mini        & EFT + PP      & NV+TG & $\checkmark$ & 4.92 & 36.10 & 10.57 & 18.08 & 106.01 & 53.48 & 45.41 & 90.59 \\
Whisper-large-v3    & EFT + PP      & NV+TG & $\checkmark$ & 17.39 & 24.34 & 10.92 & 14.86 & 154.33 & 89.06 & 384.59 & 144.68 \\
\midrule
Voxtral-Mini        & E-LoRA + PP   & NV+TG     & $\checkmark$ & 5.59 & 28.83 & 11.19 & 16.37 & 25.06 & 22.99 & 41.80 & 25.04 \\
Whisper-large-v3    & E-LoRA + PP   & NV+TG     & $\checkmark$ & 18.90 & 20.28 & 9.98 & 13.01 & 95.91 & 52.35 & 134.10 & 85.87 \\
\midrule
Voxtral-Mini        & E-LoRA + PP   & NV        & $\checkmark$ & 4.98 & 46.86 & 16.33 & 25.30 & 23.80 & 22.01 & 42.53 & 23.92 \\
Voxtral-Mini        & E-LoRA + PP   & NV+TG+CV  & $\checkmark$ & 5.71 & 30.50 & 11.30 & 16.94 & 30.84 & 21.94 & 44.59 & 28.97 \\
\bottomrule
\end{tabular}
}
\label{tab:combined_wer_full}
\end{table*}

\subsection{Baseline results}
Among open-weight models, \texttt{omniASR\_CTC\_1B\_v2} achieves the lowest WER on S-DiverSe, while Voxtral-Mini leads on NeuroVoz and Whisper-large-v3 on TORGO. No single open-weight model dominates across all corpora, suggesting that architecture and training data interact differently with each pathology type. Voxtral-Mini's sharp degradation on S-DiverSe relative to its $\sim$3\% WER on standard Spanish \cite{liu2025voxtral} reflects the inherent difficulty of pathological speech rather than language modeling deficits. \texttt{Scribe\_v2} achieves the best overall performance on S-DiverSe, consistent with exposure to broader and more diverse training data. Evaluation on TORGO and NeuroVoz was precluded by cost and NeuroVoz's license, which prohibits third-party commercial data transmission.

\begin{figure}[h]
    \centering
    \includegraphics[width=1.0\linewidth]{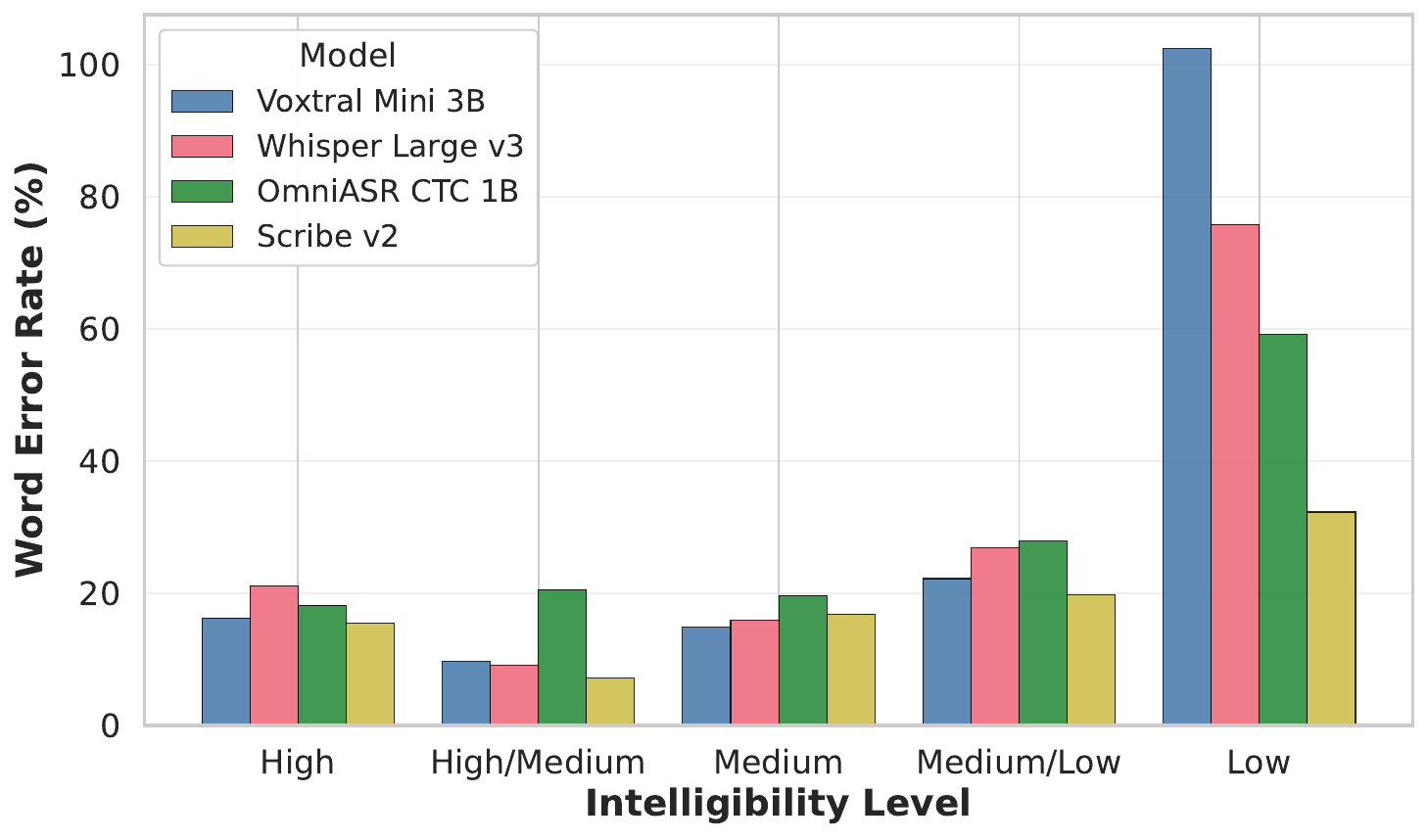}
    \vspace{-0.6cm}
    \caption{WER by intelligibility level.}
    \label{fig:wer_by_inteligibility}
\end{figure}

Figure~\ref{fig:wer_by_inteligibility} shows the expected inverse relationship between intelligibility and WER. The ``High" bin comprises PD speech exclusively, while the ``High/Medium" bin comprises uniquely ALS samples. This makes these bins not directly comparable to others. At low intelligibility, \texttt{Scribe\_v2} maintains relative robustness while open-weight models degrade substantially.

\begin{figure}[h]
    \centering
    \includegraphics[width=1.0\linewidth]{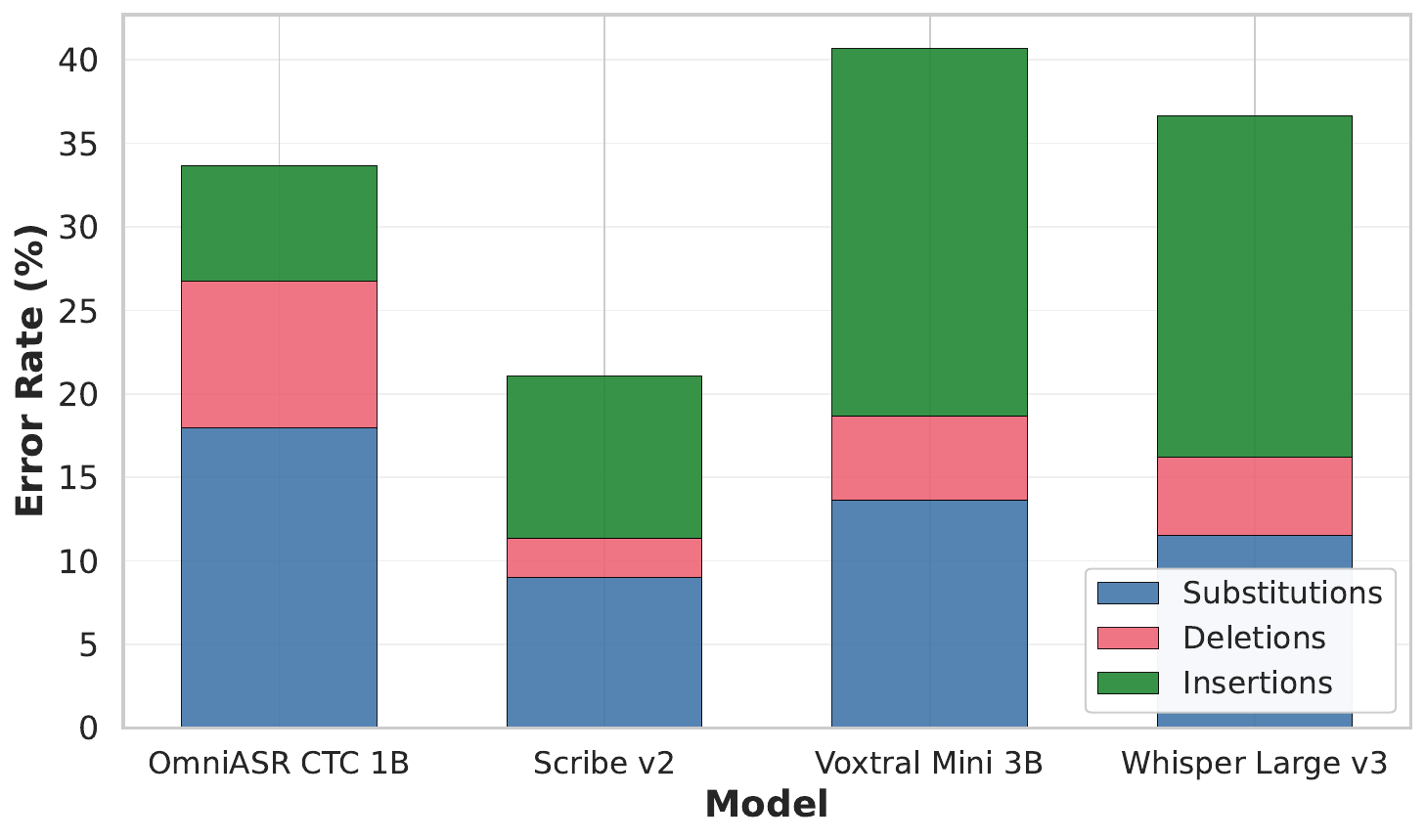}
    \vspace{-0.7cm}
    \caption{WER by ASR model decomposed by error type.}
    \label{fig:wer_and_error_type}
\end{figure}

Figure~\ref{fig:wer_and_error_type} reveals distinct error profiles across models. Autoregressive models exhibit insertion-heavy patterns consistent with hallucination under adverse acoustic conditions, a consequence of generation continuing beyond actual speech content. \texttt{omniASR\_CTC\_1B\_v2} is substitution-dominated; its frame-level token assignment inherently limits hallucinations. \texttt{Scribe\_v2} exhibits a more balanced error distribution.

\subsection{Adapted models}
PP is the most robust adaptation strategy on S-DiverSe: it consistently reduces WER for both models without harming in-domain performance on TORGO, confirming that a substantial share of errors is correctable at the text level. Fine-tuning consistently improves in-domain performance but fails to generalize to out-of-domain data.

FFT+PP yields strong in-domain gains on TORGO but severely degrades generalization to S-DiverSe, with Whisper-large-v3 collapsing to WER values exceeding 100\%, driven by an increase in insertions and substitutions under domain shift. LoRA-based methods improve generalization over FFT but do not consistently surpass PP-only on S-DiverSe. E-LoRA transfers more reliably for Voxtral-Mini than for Whisper-large-v3, suggesting that the benefit of partial parameter updates is architecture-dependent. The overall pattern is consistent with catastrophic forgetting of broadly useful representations, particularly acute for Whisper-large-v3.

To investigate the effect of language imbalance in the training set, we conducted additional experiments with Voxtral-Mini using E-LoRA+PP. Training on NeuroVoz alone yielded a PD performance that surpassed the PP-only baseline, highlighting the value of in-language pathological data despite its limited size. Complementing TORGO and NeuroVoz with Spanish Common Voice (balanced English and Spanish) nonetheless degraded S-DiverSe performance relative to NeuroVoz+TORGO alone, indicating the bottleneck is domain mismatch, not language imbalance: clean read-speech acoustics cannot bridge the gap to in-the-wild pathological speech, consistent with the perplexity analysis in Section \ref{sec:stats}.

The failure of fine-tuning to generalize to S-DiverSe reflects a genuine domain gap: in-the-wild pathological Spanish remains underrepresented in existing resources, and curating dedicated benchmarks and training data for this domain is a clear community priority.

\subsection{Limitations}
S-DiverSe relies on self-reported diagnoses and is intended for ASR evaluation rather than clinical inference. Data imbalances mirror real-world in-the-wild availability. The dataset size is modest but comparable to existing pathological speech corpora.

\section{Conclusion}

We present S-DiverSe, the first Spanish corpus of neurologically affected speech spanning multiple diseases. The dataset contains 3.2 hours of in-the-wild segments from 22 speakers with ALS, Parkinson's disease, and stroke. Our experiments reveal that current ASR systems struggle under the heterogeneous captured conditions. Heuristic text post-processing proves more robust than fine-tuning for out-of-domain neurological Spanish speech, as parameter-updating methods fail to generalize beyond the training distribution regardless of language composition. These results highlight a genuine domain gap that clean read-speech and controlled clinical corpora cannot substitute. S-DiverSe establishes a benchmark that exposes where current systems fall short, and underscores the need for dedicated data collection with broader coverage of stroke and female speakers to advance robust ASR for pathological Spanish speech.

\section{Generative AI Use Disclosure}
We used a generative AI tool to paraphrase and polish portions of the manuscript to improve readability and grammar.

\section{Acknowledgments}
\label{sec:ack}
We would like to thank Irene Gordo Bernat, speech-language pathologist, for her valuable insights and assistance with speech analysis and data selection. This project has been partially funded by the European Union’s Horizon 2020 RIA ELOQUENCE project (Grant Agreement No. 101135916). Views and opinions expressed are, however, those of the author(s) only and do not necessarily reflect those of the European Union or European Commission-EU. Neither the European Union nor the granting authority can be held responsible for them.
Santosh Kesiraju is supported by the Ministry of Education, Youth and Sports of the Czech Republic (MoE) through the OP JAK project ``Linguistics, Artificial Intelligence and Language and Speech Technologies: from Research to Applications'' (ID:CZ.02.01.01/00/23\_020/0008518).


\end{document}